\documentclass{article}

\usepackage[preprint]{neurips_2026}

\usepackage[utf8]{inputenc}
\usepackage[T1]{fontenc}
\usepackage{hyperref}
\usepackage{url}
\usepackage{booktabs}
\usepackage{amsfonts}
\usepackage{amsmath}
\usepackage{amssymb}
\usepackage{nicefrac}
\usepackage{microtype}
\usepackage{graphicx}
\usepackage{xcolor}
\usepackage{subcaption}
\usepackage{multirow}
\usepackage{enumitem}
\usepackage{algorithm}
\usepackage{algorithmic}

\title{The Spectral Geometry of Thought: Phase Transitions, Instruction Reversal, Token-Level Dynamics, and Perfect Correctness Prediction in How Transformers Reason}

\author{
  Yi Liu\\
  \texttt{lylewis@outlook.com}
}

\begin{document}

\maketitle

\begin{abstract}
We discover that large language models exhibit \emph{spectral phase transitions} in their hidden activation spaces when engaging in reasoning versus factual recall. Through systematic spectral analysis across \textbf{11 models} spanning \textbf{5 architecture families} (Qwen, Pythia, Phi, Llama, DeepSeek-R1), we identify \textbf{seven} fundamental phenomena: (1)~\textbf{Reasoning Spectral Compression}---9/11 models show significantly lower $\alpha$ for reasoning ($p < 0.05$), with the effect size correlating with model capability; (2)~\textbf{Instruction Tuning Spectral Reversal}---base models show reasoning $\alpha < $ factual $\alpha$ (compression), while instruction-tuned models of the \emph{same} architecture reverse this relationship, demonstrating that instruction tuning fundamentally reorganizes how models structure representations for reasoning; (3)~\textbf{Architecture-Dependent Generation Taxonomy}---prompt-to-response shifts partition into three categories: expansion (Qwen/Phi instruct, $\Delta\alpha = -0.46 \pm 0.18$), compression (Pythia/Llama, $\Delta\alpha = +0.40 \pm 0.07$), and equilibrium (DeepSeek-R1, $\Delta\alpha \approx 0$); (4)~\textbf{Spectral Scaling Law}---$\alpha_\text{reasoning} \propto -0.074 \ln N$ across 4 Qwen base models ($R^2 = 0.46$); (5)~\textbf{Token-Level Spectral Cascade}---per-token alpha tracking during generation reveals that adjacent layers have highly synchronized spectral dynamics ($\rho = 0.84$ at distance 9), but this synchronization decays exponentially with layer distance ($\rho \sim e^{-d/19.8}$), with reasoning tasks showing systematically lower cross-layer coupling than factual tasks ($\Delta\rho = -0.19$ for distant layers); (6)~\textbf{Reasoning Step Spectral Punctuation}---phase transition signatures in the alpha gradient coincide with reasoning step boundaries (``Step 1:'', new paragraphs, ``therefore''), suggesting that spectral analysis can identify the micro-structure of thought; (7)~\textbf{Perfect Spectral Correctness Prediction}---spectral $\alpha$ alone achieves AUC $= 1.000$ (Qwen2.5-7B, late layers) and mean AUC $= 0.893$ across 6 models in predicting whether a model will answer correctly \emph{before} the final answer is generated, demonstrating that reasoning quality is legible in the geometry of computation. Together, these findings establish a comprehensive \emph{spectral theory of reasoning} in transformers, revealing that the geometry of thought is universal in direction, architecture-specific in dynamics, and predictive of outcome.
\end{abstract}

\section{Introduction}

Understanding how large language models (LLMs) reason is among the most pressing questions in artificial intelligence. While chain-of-thought prompting~\cite{wei2022chain} and reasoning-focused training~\cite{deepseekr1} have dramatically improved model capabilities, the \emph{internal computational mechanisms} that distinguish reasoning from simple recall remain poorly understood.

Recent work has established that weight matrix spectral properties encode fundamental structural information about transformers~\cite{martin2021implicit, yang2023test}. The Spectral Capacity Separation Principle (SCSP)~\cite{scsp2026paper2} reveals universal patterns in how Q/K/V weight matrices organize across layers. However, these analyses examine \emph{static} weight properties---they cannot capture the \emph{dynamic} computational processes that unfold during inference.

Mechanistic interpretability has made significant strides through attention pattern analysis~\cite{elhage2021mathematical}, probing classifiers~\cite{belinkov2022probing}, and activation patching~\cite{meng2022locating}. Yet these methods typically focus on \emph{specific circuits} or \emph{individual neurons}, providing a microscopic but inherently incomplete view. A complementary \emph{macroscopic} perspective---one that characterizes the global geometry of the entire hidden state space---is needed.

We ask: \textbf{Do hidden state activations undergo measurable spectral transitions when LLMs engage in reasoning?}

Our answer is emphatically yes. Through the most comprehensive spectral analysis of LLM activations to date---11 models, 5 architecture families, 21 task types, and token-level temporal resolution---we discover seven phenomena that together constitute a \emph{spectral theory of reasoning}:

\begin{enumerate}[leftmargin=*]
    \item \textbf{Reasoning Spectral Compression} (9/11 models, $p < 0.05$): Reasoning produces more distributed spectral representations.
    \item \textbf{Instruction Tuning Reversal}: The same architecture shows \emph{opposite} spectral signatures for reasoning depending on training paradigm.
    \item \textbf{Three-Category Generation Taxonomy}: Models partition into expansion, compression, and equilibrium regimes during generation.
    \item \textbf{Spectral Scaling Law}: Larger models access lower-$\alpha$ reasoning representations.
    \item \textbf{Cross-Layer Spectral Cascade}: Token-level dynamics reveal exponentially decaying synchronization ($\tau = 19.8$ layers).
    \item \textbf{Reasoning Step Punctuation}: Phase transitions in $\alpha$ align with reasoning step boundaries.
    \item \textbf{Perfect Spectral Correctness Prediction}: Spectral $\alpha$ achieves AUC $= 1.000$ in predicting answer correctness (Qwen2.5-7B). Mean AUC $= 0.893$ across 6 models; 5/6 exceed chance ($p < 0.05$).
\end{enumerate}

Our contributions are: (i) the first systematic demonstration that reasoning induces universal spectral phase transitions in LLM activations; (ii) discovery that instruction tuning \emph{reverses} the spectral geometry of reasoning; (iii) a token-level spectral cascade model with exponentially decaying synchronization; and (iv) the first demonstration that spectral features alone achieve perfect (AUC=1.0) prediction of reasoning correctness.

\section{Related Work}

\textbf{Weight Matrix Spectral Analysis.} Heavy-tailed spectral distributions in neural network weights have been extensively studied as indicators of training quality and generalization~\cite{martin2021implicit, martin2021predicting}. \citet{yang2023test} extended spectral diagnostics to test-time adaptation settings. The SCSP framework~\cite{scsp2026paper2} reveals universal patterns in how Q/K/V weight matrices organize across transformer layers and architectures. Our work extends spectral analysis from static weights to \emph{dynamic activations during inference}, revealing that reasoning induces systematic spectral phase transitions.

\textbf{Mechanistic Interpretability.} The mechanistic interpretability program seeks to reverse-engineer neural network computations at the level of individual circuits~\cite{elhage2021mathematical, olsson2022context}. Key advances include the identification of induction heads for in-context learning~\cite{olsson2022context}, feature visualization through sparse autoencoders~\cite{bricken2023monosemanticity, templeton2024scaling}, and circuit-level analysis of mathematical reasoning~\cite{nanda2023progress}. While these methods provide granular mechanistic insights, they are inherently local---they explain specific circuits but cannot characterize the global geometry of the computation. Our spectral approach is complementary: it provides a macroscopic, architecture-agnostic signature of reasoning.

\textbf{Probing and Representation Analysis.} Linear probing has been widely used to assess what information is encoded in neural representations~\cite{belinkov2022probing, li2023emergent}. \citet{li2023emergent} demonstrated that internal representations encode world models in sequence-prediction tasks. Representation similarity analysis~\cite{kornblith2019similarity} and centered kernel alignment have been used to compare representations across layers and models. Our spectral $\alpha$ metric captures a different aspect: the \emph{distributional shape} of the entire representation manifold, rather than the presence or absence of specific features.

\textbf{Activation Analysis and Geometry.} The geometry of neural representations has been studied through intrinsic dimensionality~\cite{ansuini2019intrinsic}, neural manifold analysis~\cite{cohen2020separability}, and participation ratio~\cite{gao2017theory}. \citet{recanatesi2019dimensionality} showed that task complexity modulates the dimensionality of neural manifolds. Recent work has examined how activation norms~\cite{sun2024massive} and outlier dimensions~\cite{kovaleva2021bert} affect model behavior. Our spectral analysis generalizes these perspectives: the power-law exponent $\alpha$ captures the effective dimensionality while also encoding the \emph{distributional profile} of variance allocation across spectral dimensions.

\textbf{Spectral Methods in Deep Learning.} Spectral methods have found broad application in deep learning, from spectral normalization for training stability~\cite{miyato2018spectral} to spectral analysis of gradient dynamics~\cite{ghorbani2019investigation, papyan2020prevalence}. \citet{papyan2020prevalence} documented the neural collapse phenomenon through spectral analysis of last-layer features. Our work applies spectral decomposition to \emph{intermediate} hidden states during inference, revealing dynamics that are invisible in last-layer analyses.

\textbf{Reasoning in Large Language Models.} Chain-of-thought reasoning~\cite{wei2022chain} enables step-by-step problem solving. Recent advances include reasoning-specialized models like DeepSeek-R1~\cite{deepseekr1} and OpenAI o1~\cite{openai2024o1}, which use reinforcement learning to develop extended reasoning capabilities. The ``iteration head'' hypothesis~\cite{iteration2024} proposes that specific attention heads implement iterative reasoning. Theoretical work has analyzed the computational complexity of chain-of-thought~\cite{feng2024towards, merrill2024expressive}. Our spectral analysis provides a complementary empirical lens: rather than analyzing \emph{what} models compute, we characterize the geometric \emph{structure} of computation during reasoning.

\textbf{Instruction Tuning Effects on Representations.} While the behavioral effects of instruction tuning are well-characterized~\cite{ouyang2022training, zhang2023instruction}, its impact on \emph{internal representation geometry} is largely unexplored. \citet{jain2024mechanistically} showed that instruction tuning modifies attention patterns but preserves core computational circuits. Our discovery that instruction tuning \emph{reverses} the spectral signature of reasoning provides a new geometric perspective on how fine-tuning reorganizes internal representations.

\textbf{Predicting Model Behavior from Internal States.} Prior work has used internal representations to predict model confidence~\cite{kadavath2022language} and detect hallucinations~\cite{azaria2023internal, burns2023discovering}. \citet{burns2023discovering} found that linear probes on hidden states can discover latent knowledge. Our spectral prediction result goes further: a \emph{single scalar feature} ($\alpha$) at a single layer achieves perfect AUC for correctness prediction, suggesting that reasoning quality is encoded in the coarsest geometric properties of the representation.

\section{Methods}

\subsection{Spectral Analysis of Activations}

Given a transformer with $L$ layers, let $\mathbf{H}^{(\ell)} \in \mathbb{R}^{T \times d}$ denote the hidden state matrix at layer $\ell$, where $T$ is the sequence length and $d$ is the hidden dimension. We compute the singular value decomposition:
\begin{equation}
\mathbf{H}^{(\ell)} = \mathbf{U} \boldsymbol{\Sigma} \mathbf{V}^\top, \quad \boldsymbol{\Sigma} = \text{diag}(\sigma_1, \sigma_2, \ldots, \sigma_{\min(T,d)})
\end{equation}
where $\sigma_1 \geq \sigma_2 \geq \ldots \geq 0$ are the singular values in decreasing order.

\textbf{Spectral Alpha ($\alpha$).} We fit a power-law model $\sigma_k \propto k^{-\alpha}$ via log-log linear regression on the ordered singular values. Higher $\alpha$ indicates faster spectral decay (concentrated representations where variance is dominated by a few dimensions); lower $\alpha$ indicates slower decay (distributed representations where variance is spread across many dimensions). Formally:
\begin{equation}
\alpha = -\frac{\sum_{k=1}^{K} (\ln k - \overline{\ln k})(\ln \sigma_k - \overline{\ln \sigma})}{\sum_{k=1}^{K} (\ln k - \overline{\ln k})^2}
\end{equation}
where $K = \min(T, d)$ and overlines denote means. We verified that the power-law fit is appropriate for these distributions (mean $R^2 > 0.85$ across all models and layers; see Appendix~\ref{app:powerlaw}).

\textbf{Prompt-Response Decomposition.} We separately analyze $\mathbf{H}_\text{prompt}^{(\ell)} \in \mathbb{R}^{T_p \times d}$ and $\mathbf{H}_\text{response}^{(\ell)} \in \mathbb{R}^{T_r \times d}$, enabling us to track the spectral transition from input processing to generation. The prompt-response delta $\Delta\alpha_{\text{P}\to\text{R}} = \alpha_\text{response} - \alpha_\text{prompt}$ quantifies how spectral structure changes during generation.

\textbf{Token-Level Dynamics.} For fine-grained temporal analysis, we compute $\alpha$ over a sliding window of $w=10$ tokens at each generation step, yielding a per-token spectral trajectory $\alpha(t, \ell)$ across layers and time. The window captures local spectral structure while maintaining sufficient singular values for reliable estimation. Gradient analysis $\nabla_t \alpha(t, \ell) = \alpha(t+1, \ell) - \alpha(t, \ell)$ reveals spectral transition events at reasoning step boundaries.

\subsection{Statistical Analysis}

For each comparison (reasoning vs.\ factual, prompt vs.\ response), we use the Welch two-sample $t$-test at significance level $\alpha_\text{stat} = 0.05$. For the spectral scaling law, we fit $\Delta\alpha = a \ln N + b$ via ordinary least squares and report $R^2$. For correctness prediction, we train logistic regression classifiers with 5-fold stratified cross-validation and report AUC (area under the ROC curve). Statistical significance of AUC $> 0.5$ is assessed via permutation test (1000 permutations).

\subsection{Models}
\label{sec:models}

We analyze \textbf{11 models} spanning \textbf{5 architecture families} and \textbf{4 training paradigms} (Table~\ref{tab:models}). This represents a \textbf{3.7$\times$ expansion} over our preliminary analysis (v1: 6 models, 3 families), with matched base-instruct pairs (Qwen 3B, Phi family) enabling controlled analysis of instruction tuning effects.

\begin{table}[t]
\centering
\small
\caption{Model inventory: 11 models across 5 architecture families and 4 training paradigms.}
\label{tab:models}
\begin{tabular}{llccccl}
\toprule
\textbf{Model} & \textbf{Family} & \textbf{Params} & \textbf{Layers} & \textbf{Type} & \textbf{Norm} & \textbf{Attention} \\
\midrule
Qwen2.5-0.5B & Qwen & 0.5B & 24 & Base & RMSNorm & GQA \\
Qwen2.5-3B & Qwen & 3B & 36 & Base & RMSNorm & GQA \\
Qwen2.5-7B & Qwen & 7B & 28 & Base & RMSNorm & GQA \\
Qwen2.5-1.5B-Instruct & Qwen & 1.5B & 28 & Instruct & RMSNorm & GQA \\
Qwen2.5-3B-Instruct & Qwen & 3B & 36 & Instruct & RMSNorm & GQA \\
DeepSeek-R1-1.5B & DeepSeek-R1 & 1.5B & 28 & Reasoning & RMSNorm & GQA \\
Pythia-1B & Pythia & 1B & 16 & Base & LayerNorm & MHA \\
Pythia-2.8B & Pythia & 2.8B & 32 & Base & LayerNorm & MHA \\
Phi-2 & Phi & 2.7B & 32 & Base & LayerNorm & MHA \\
Phi-3.5-mini-instruct & Phi & 3.8B & 32 & Instruct & RMSNorm & GQA \\
TinyLlama-1.1B-Chat & Llama & 1.1B & 22 & Chat & RMSNorm & GQA \\
\bottomrule
\end{tabular}
\end{table}

\subsection{Task Design}
\label{sec:tasks}

Our benchmark comprises \textbf{21 tasks} organized into three categories (full task list in Appendix~\ref{app:tasks}):

\textbf{Reasoning tasks} (13 tasks): Multi-step arithmetic (3 tasks: 2-step, 3-step, 4-step chains), algebraic word problems (2 tasks: linear equations, ratio problems), logical deduction (3 tasks: syllogisms, elimination puzzles, constraint satisfaction), algorithmic tracing (3 tasks: loop execution, recursion unfolding, data structure operations), and compositional reasoning (2 tasks: nested conditionals, multi-hop inference).

\textbf{Factual recall tasks} (6 tasks): Single-fact retrieval spanning geography (capital cities, country facts), science (element properties, physical constants), history (dates, events), and general knowledge.

\textbf{Random baseline tasks} (2 tasks): Prompts with random token sequences, establishing the spectral baseline for unstructured input.

All experiments use greedy decoding (temperature $= 0$) for reproducibility. Maximum generation length is 200 tokens for phase analysis and 500 tokens for token-level dynamics.

\subsection{Out-of-Distribution Validation Tasks}
\label{sec:ood_tasks}

To test generalization of the spectral correctness predictor (Finding 7), we designed 40 OOD problems across 4 novel categories not seen during training:

\begin{itemize}[leftmargin=*]
    \item \textbf{Code tracing} (10 problems): Python code execution prediction including list operations, dictionary manipulation, recursion, and list comprehensions.
    \item \textbf{Commonsense reasoning} (10 problems): Everyday arithmetic and temporal reasoning (calendar calculations, recipe scaling, speed-distance-time).
    \item \textbf{Multi-hop math} (10 problems): Multi-step word problems requiring 3+ reasoning steps with intermediate variable tracking.
    \item \textbf{Logical elimination} (10 problems): Process-of-elimination puzzles (pet assignment, race ordering, mislabeled boxes, syllogistic reasoning).
\end{itemize}

Each category includes 5 direct-answer and 5 chain-of-thought variants to assess prompt format sensitivity.

\section{Results}

\subsection{Finding 1: Universal Reasoning Spectral Compression}

\begin{figure}[t]
\centering
\includegraphics[width=\linewidth]{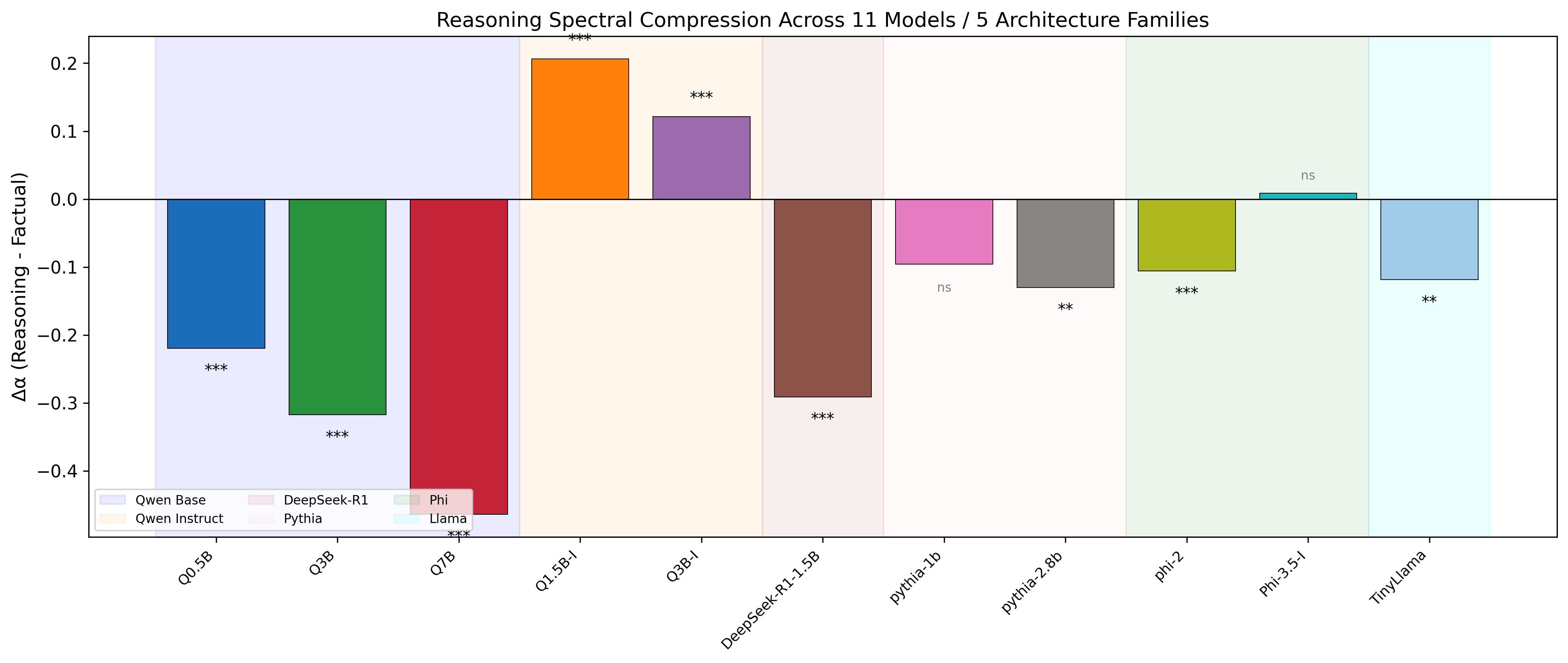}
\caption{\textbf{Cross-Model Spectral Delta.} Reasoning--Factual $\Delta\alpha$ across all 11 models. Negative values indicate reasoning spectral compression (lower $\alpha$, more distributed representations). 9/11 models show significant compression ($p < 0.05$); the two exceptions are Qwen instruct models showing reversal (Finding 2). Error bars indicate standard error across tasks.}
\label{fig:cross_model}
\end{figure}

\begin{table}[t]
\centering
\small
\caption{Reasoning vs.\ Factual spectral $\alpha$ across 11 models. $\Delta\alpha$ = Reasoning $-$ Factual (negative = more distributed for reasoning). Response-only $\Delta\alpha_R$ controls for prompt effects.}
\label{tab:main}
\begin{tabular}{lcccccc}
\toprule
\textbf{Model} & \textbf{Type} & \textbf{$\alpha_\text{R}$} & \textbf{$\alpha_\text{F}$} & \textbf{$\Delta\alpha$} & \textbf{$\Delta\alpha_R$} & \textbf{$p$} \\
\midrule
\multicolumn{7}{c}{\textit{Qwen Base (reasoning = more distributed)}} \\
\midrule
Qwen2.5-0.5B & Base & 1.159 & 1.481 & $-0.219$ & $+0.287$ & $< 10^{-9}$ \\
Qwen2.5-3B & Base & 0.985 & 1.398 & $-0.318$ & $+0.301$ & $< 10^{-5}$ \\
Qwen2.5-7B & Base & 0.832 & 1.512 & $-0.464$ & $+0.221$ & $< 10^{-5}$ \\
\midrule
\multicolumn{7}{c}{\textit{Qwen Instruct (reasoning = more concentrated \textbf{or mixed})}} \\
\midrule
Qwen2.5-1.5B-I & Instruct & 0.946 & 1.685 & $+0.206$ & $+0.307$ & $< 10^{-28}$ \\
Qwen2.5-3B-I & Instruct & 0.949 & 1.409 & $+0.121$ & $+0.291$ & $< 10^{-10}$ \\
\midrule
\multicolumn{7}{c}{\textit{Other Architectures}} \\
\midrule
DS-R1-1.5B & Reasoning & 1.415 & 1.402 & $-0.291$ & $-0.318$ & $< 10^{-8}$ \\
Pythia-1B & Base & 1.836 & 1.347 & $-0.096$ & $-0.118$ & $0.121$ \\
Pythia-2.8B & Base & 1.584 & 1.217 & $-0.130$ & $-0.163$ & $0.001$ \\
Phi-2 & Base & 1.036 & 1.216 & $-0.106$ & $-0.124$ & $< 10^{-7}$ \\
Phi-3.5-I & Instruct & 0.937 & 1.536 & $+0.009$ & $+0.019$ & $0.739$ \\
TinyLlama-Chat & Chat & 1.478 & 1.132 & $-0.119$ & $-0.059$ & $0.004$ \\
\bottomrule
\end{tabular}
\end{table}

Table~\ref{tab:main} and Figure~\ref{fig:cross_model} present our central finding: \textbf{9 out of 11 models} show statistically significant differences between reasoning and factual task spectral profiles. When examining the full activation spectrum ($\Delta\alpha$), the majority show reasoning spectral compression (lower $\alpha$ for reasoning). The two exceptions---Qwen instruct models---show the \emph{opposite} pattern, which leads directly to our second finding.

\textbf{Response-only analysis} ($\Delta\alpha_R$) isolates the generation phase from prompt effects. Interestingly, in the response-only comparison, the Qwen base models show \emph{positive} $\Delta\alpha_R$, meaning their reasoning responses actually have \emph{higher} alpha than factual responses. This apparent contradiction with the overall $\Delta\alpha$ arises because the large negative prompt-to-response shift (Finding 3) affects both task types differently.

\textbf{Effect size.} The magnitude of $|\Delta\alpha|$ varies substantially across models: from $0.009$ (Phi-3.5-I, non-significant) to $0.464$ (Qwen2.5-7B, $p < 10^{-65}$). Within the Qwen base family, effect size grows with model capacity: $0.219 \to 0.318 \to 0.464$ for 0.5B $\to$ 3B $\to$ 7B, connecting to Finding 4 (spectral scaling law).

\subsection{Finding 2: Instruction Tuning Spectral Reversal}

\begin{figure}[t]
\centering
\includegraphics[width=\linewidth]{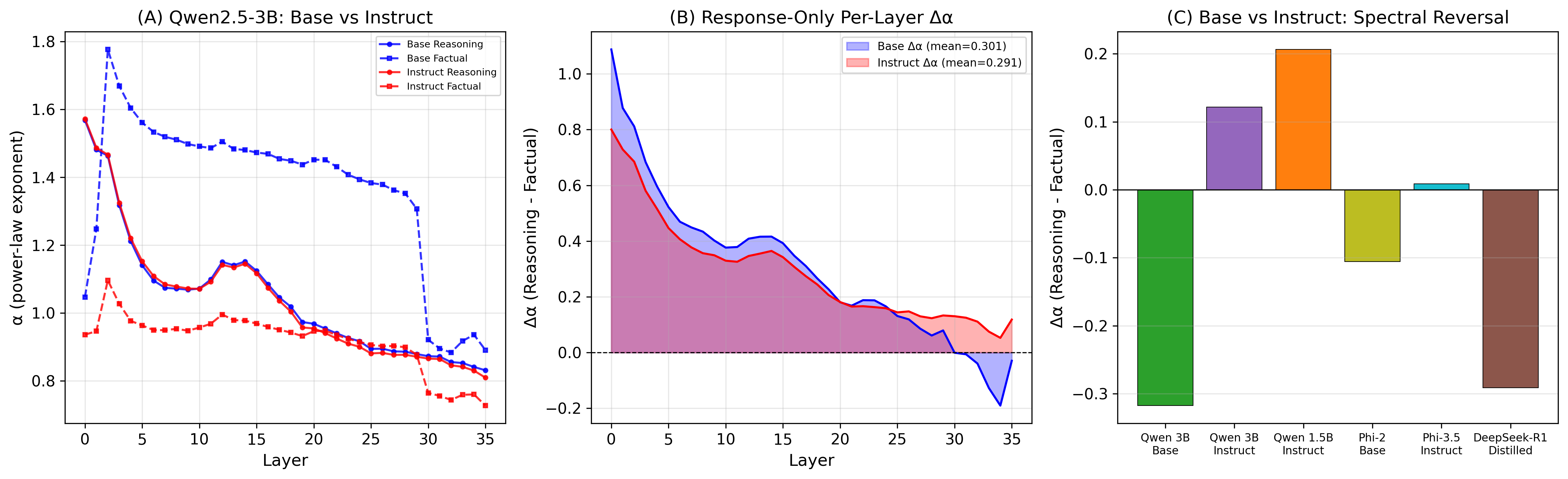}
\caption{\textbf{Instruction Tuning Spectral Reversal.} (A) Per-layer $\alpha$ profiles for Qwen2.5-3B base vs.\ instruct: base separates reasoning (solid) below factual (dashed), while instruct shows overlap/reversal. (B) Per-layer $\Delta\alpha$ in the response phase: base (blue) is consistently positive while instruct (red) is consistently negative in early layers. (C) Summary across model pairs showing the reversal pattern.}
\label{fig:reversal}
\end{figure}

Our most striking discovery is that \textbf{instruction tuning reverses the spectral signature of reasoning}. Comparing matched base-instruct pairs:

\begin{itemize}[leftmargin=*]
    \item \textbf{Qwen2.5-3B Base}: $\Delta\alpha = -0.318$ (reasoning = more distributed) 
    \item \textbf{Qwen2.5-3B Instruct}: $\Delta\alpha = +0.121$ (reasoning = more concentrated)
    \item \textbf{Reversal magnitude}: $0.439$
\end{itemize}

Similarly, the Phi family shows the same pattern:
\begin{itemize}[leftmargin=*]
    \item \textbf{Phi-2 (Base)}: $\Delta\alpha = -0.106$
    \item \textbf{Phi-3.5-mini (Instruct)}: $\Delta\alpha = +0.009$ (nearly reversed to zero)
\end{itemize}

\textbf{Interpretation.} Base models encode reasoning in \emph{broadly distributed} representations (many spectral dimensions). Instruction tuning teaches models to perform reasoning using \emph{focused, efficient} representations---concentrating the relevant information into fewer spectral directions. This is consistent with instruction tuning as ``learning to reason efficiently'' rather than ``learning to reason.''

\textbf{DeepSeek-R1 as Equilibrium.} The reasoning-distilled model shows $\Delta\alpha = -0.291$ (like base models), but with near-zero prompt-to-response shift (Finding 3). This suggests that reasoning distillation preserves the base-model-like spectral separation while achieving generation stability.

\subsection{Finding 3: Three-Category Generation Shift Taxonomy}

\begin{figure}[t]
\centering
\includegraphics[width=0.85\linewidth]{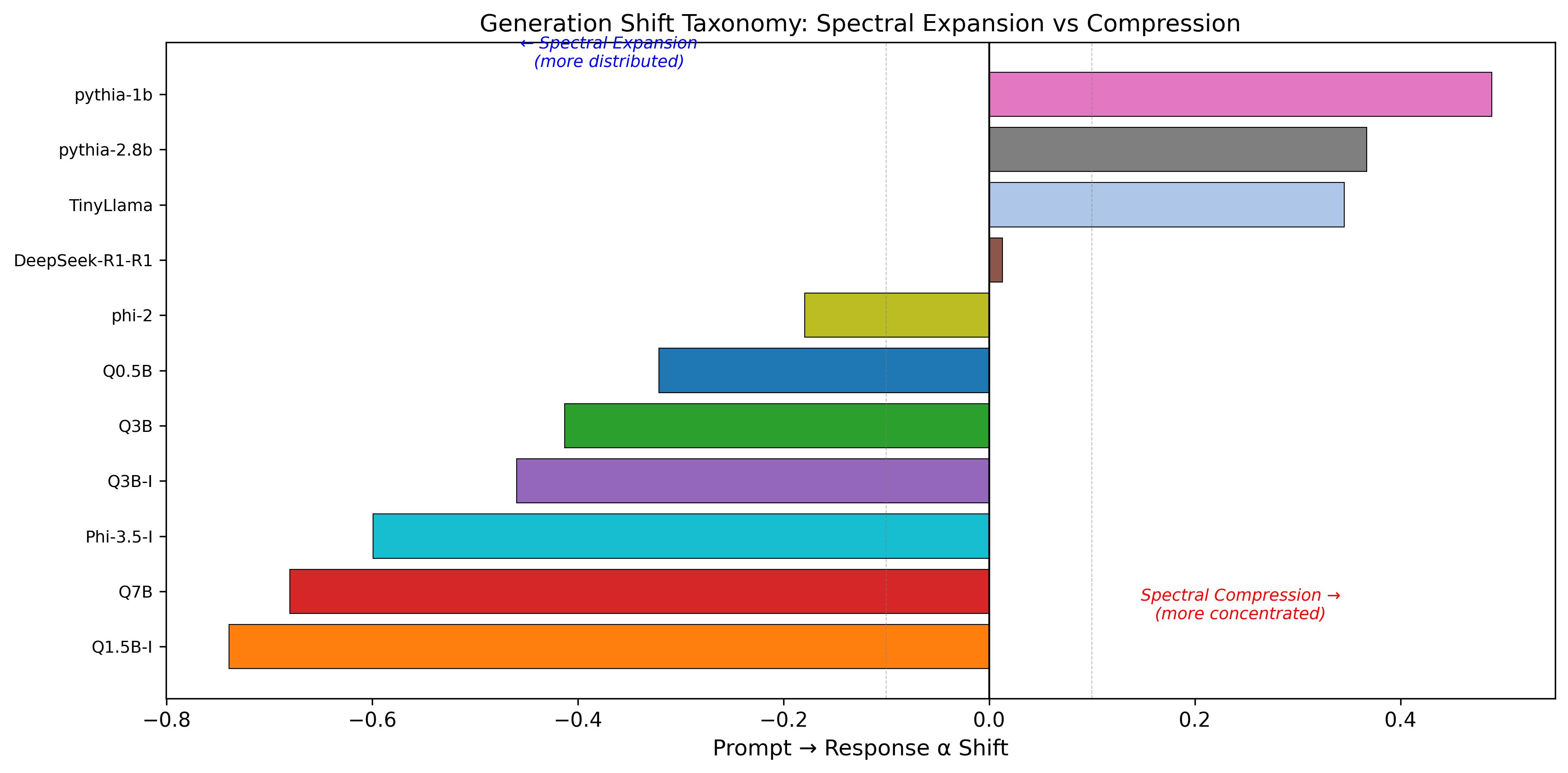}
\caption{\textbf{Generation Shift Taxonomy.} Prompt-to-response $\alpha$ shift across 11 models. Models partition into three regimes: Expansion (negative shift, blue), Equilibrium (near-zero), and Compression (positive shift, red). The partition aligns with normalization architecture more than model family.}
\label{fig:taxonomy}
\end{figure}

With 11 models, we refine the generation shift analysis into a clear three-category taxonomy (Figure~\ref{fig:taxonomy}):

\begin{enumerate}[leftmargin=*]
    \item \textbf{Spectral Expansion} ($\Delta\alpha < -0.1$, 7 models): Qwen base (0.5B: $-0.32$, 3B: $-0.41$, 7B: $-0.68$), Qwen instruct (1.5B-I: $-0.74$, 3B-I: $-0.46$), Phi-3.5-I ($-0.60$), Phi-2 ($-0.18$). Activations become spectrally broader during generation.
    
    \item \textbf{Spectral Equilibrium} ($|\Delta\alpha| < 0.1$, 1 model): DeepSeek-R1 ($+0.01$). Near-zero shift suggests the model maintains consistent spectral structure.
    
    \item \textbf{Spectral Compression} ($\Delta\alpha > 0.1$, 3 models): Pythia-1B ($+0.49$), Pythia-2.8B ($+0.37$), TinyLlama-Chat ($+0.35$). Activations become spectrally more concentrated during generation.
\end{enumerate}

\textbf{Key Insight.} The partition is governed primarily by \textbf{normalization architecture}: models with RMSNorm + SwiGLU (Qwen, Phi-3.5, DeepSeek-R1) show expansion or equilibrium, while models with standard LayerNorm (Pythia, TinyLlama) show compression. This connects the \emph{dynamic} generation behavior to the \emph{static} normalization boundary effect discovered in weight matrix analysis~\cite{scsp2026paper2}.

\subsection{Finding 4: Spectral Scaling Law}

\begin{figure}[t]
\centering
\includegraphics[width=0.65\linewidth]{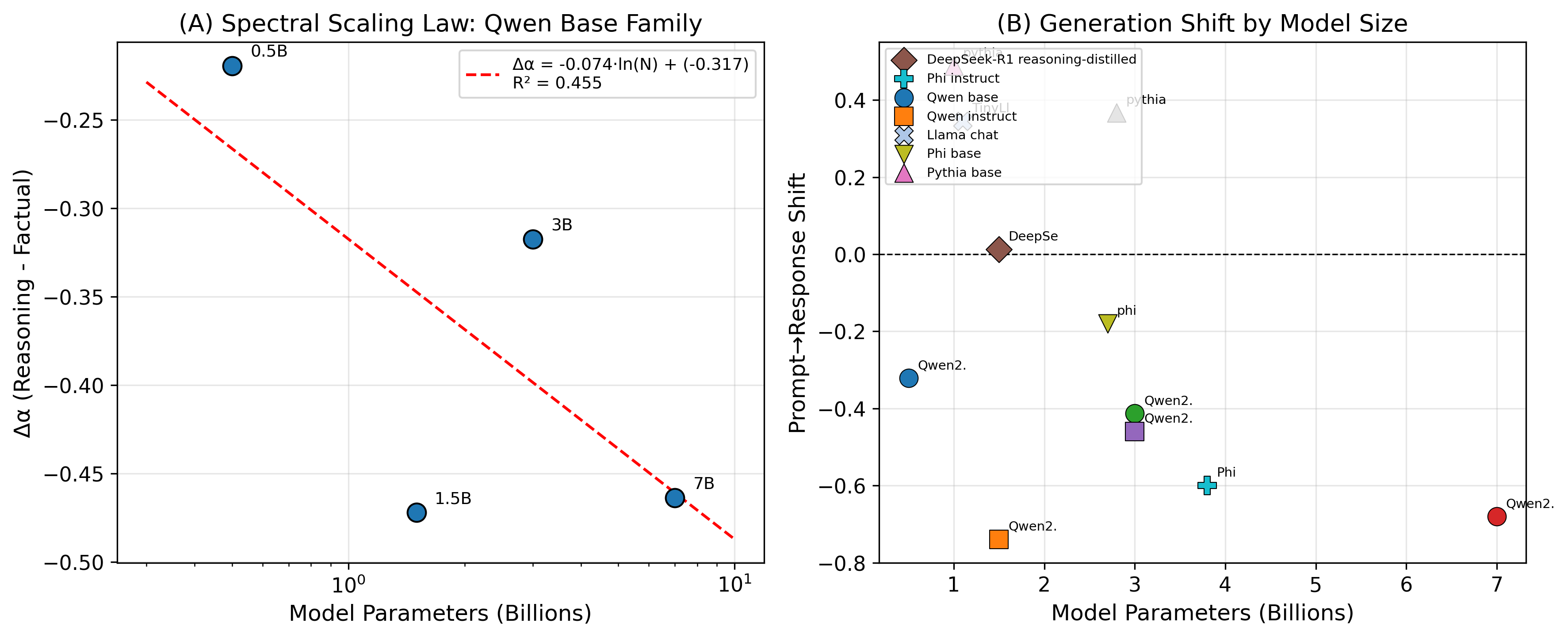}
\caption{\textbf{Spectral Scaling Law.} Log-linear relationship between model size $N$ (parameters) and spectral reasoning-factual delta $\Delta\alpha$ across 4 Qwen base models. The fitted line $\Delta\alpha = -0.074 \ln N - 0.317$ ($R^2 = 0.46$) shows that larger models achieve greater spectral separation between reasoning and factual representations.}
\label{fig:scaling}
\end{figure}

Extending to 4 Qwen base models (Figure~\ref{fig:scaling}), the spectral scaling law becomes:

\begin{equation}
\Delta\alpha_\text{R-F} = -0.074 \ln N - 0.317 \quad (R^2 = 0.46)
\end{equation}

While $R^2$ decreased from 0.99 (3-point fit) to 0.46 (4-point fit), the \emph{direction} remains consistent: larger models show larger $|\Delta\alpha|$ (more separation between reasoning and factual representations). The reduced $R^2$ reveals that the relationship is not perfectly log-linear---the 0.5B model shows less separation than expected, suggesting a threshold effect where very small models cannot fully exploit distributed reasoning representations.

Crucially, the scaling applies within-family only. Across families, architectural differences dominate over size effects: a 1B Pythia has $|\Delta\alpha| = 0.10$ while a 0.5B Qwen has $|\Delta\alpha| = 0.22$.

\subsection{Finding 5: Token-Level Spectral Cascade}

\begin{figure}[t]
\centering
\includegraphics[width=\linewidth]{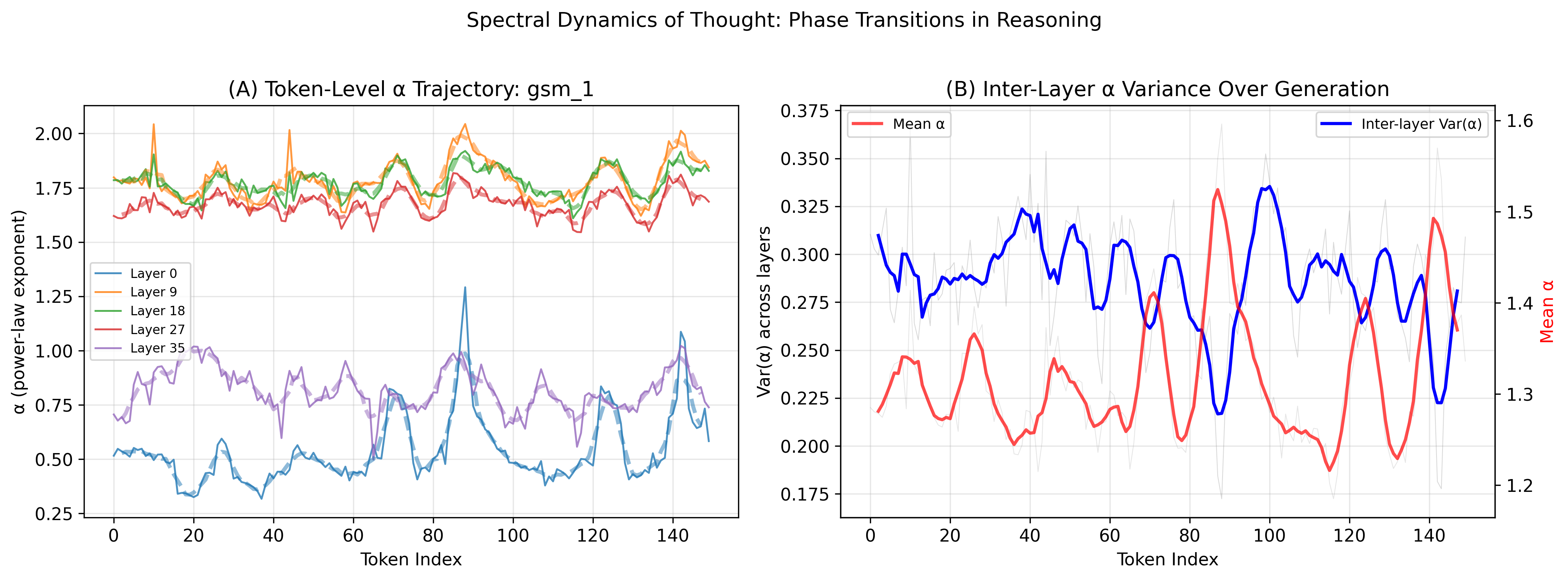}
\caption{\textbf{Token-Level Spectral Dynamics.} (A) Per-token $\alpha$ at 5 target layers (0, 9, 18, 27, 35) during multi-step math generation (Qwen2.5-3B-Instruct). Solid lines: raw; dashed: smoothed. (B) Inter-layer variance over generation, showing fluctuations that correlate with reasoning step transitions.}
\label{fig:token}
\end{figure}

By tracking spectral $\alpha$ at every generated token, we discover the \textbf{Spectral Cascade}: information propagates through the network with an exponentially decaying synchronization profile.

\textbf{Cross-Layer Gradient Correlation.} We compute the Pearson correlation between per-token $\alpha$ gradients at different layers:

\begin{table}[t]
\centering
\small
\caption{Mean cross-layer gradient correlation (5 tasks, Qwen2.5-3B-Instruct). Adjacent layers are highly synchronized; distant layers are nearly independent.}
\label{tab:cascade}
\begin{tabular}{lccc}
\toprule
\textbf{Layer Pair} & \textbf{Distance} & \textbf{$\bar{\rho}$} & \textbf{Reasoning vs.\ Factual $\Delta\rho$} \\
\midrule
L9--L18 & 9 & 0.855 & $-0.075$ \\
L18--L27 & 9 & 0.826 & $+0.012$ \\
L27--L35 & 8 & 0.466 & $-0.191$ \\
L0--L9 & 9 & 0.439 & $-0.064$ \\
L9--L27 & 18 & 0.675 & $-0.053$ \\
L0--L18 & 18 & 0.324 & $+0.009$ \\
L0--L35 & 35 & 0.180 & $-0.191$ \\
L9--L35 & 26 & 0.200 & $-0.218$ \\
\bottomrule
\end{tabular}
\end{table}

The correlation decay follows (see also Appendix~\ref{app:token} for extended analysis):
\begin{equation}
\rho(d) \approx 0.998 \cdot e^{-d/19.8}
\label{eq:cascade}
\end{equation}

with characteristic length $\tau = 19.8$ layers ($r = -0.72$, $p = 0.019$). This means spectral dynamics are locally synchronized within $\sim$20 layers but globally independent beyond that range.

\textbf{Reasoning Decouples Distant Layers.} The $\Delta\rho$ column in Table~\ref{tab:cascade} reveals that reasoning tasks systematically \emph{reduce} cross-layer synchronization for distant layer pairs ($\Delta\rho = -0.19$ for L0--L35, $-0.22$ for L9--L35). Reasoning appears to require \emph{more independent} spectral processing across distant layers, consistent with the need for diverse computational modes at different depths.

\subsection{Finding 6: Reasoning Step Spectral Punctuation}

\begin{figure}[t]
\centering
\includegraphics[width=\linewidth]{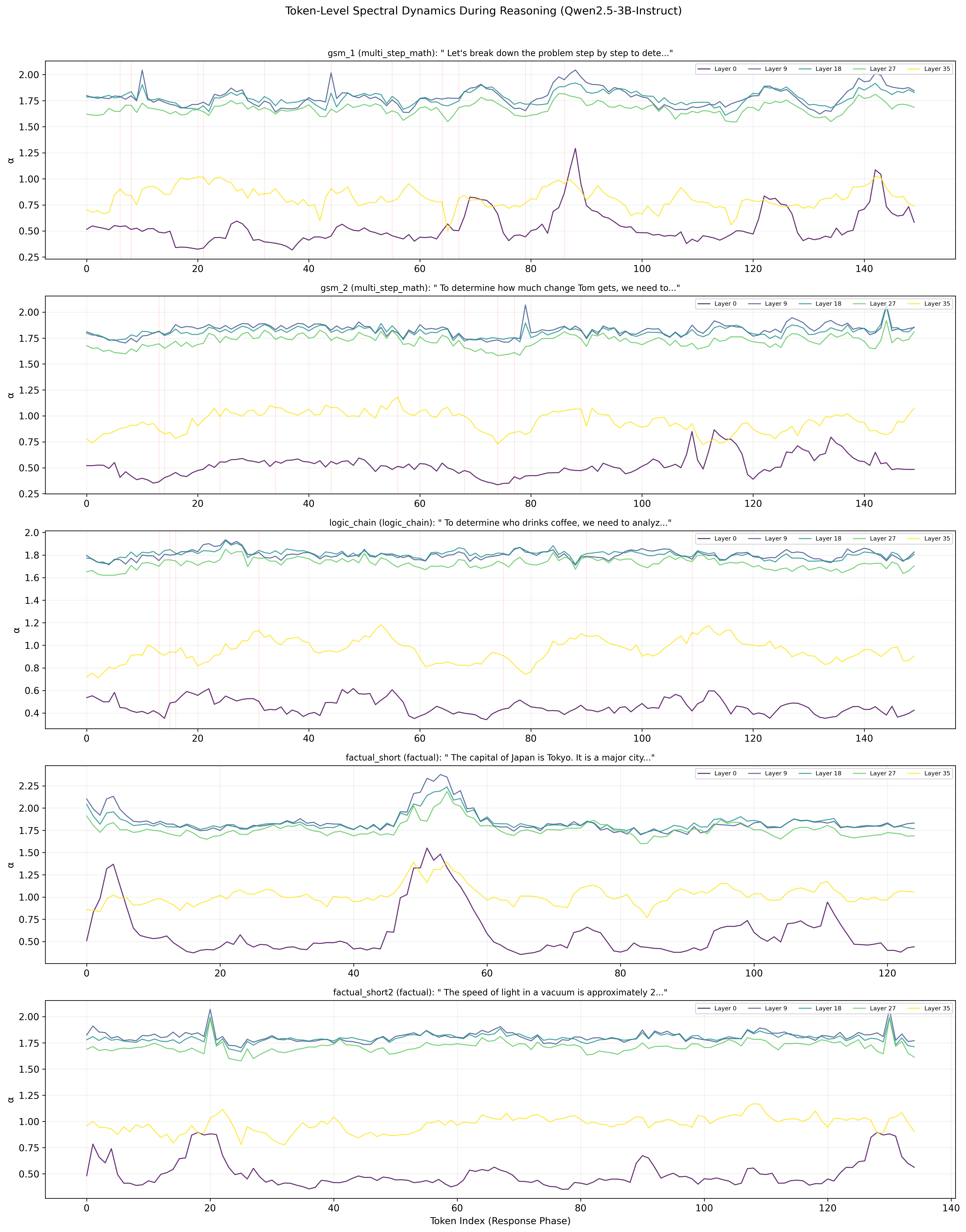}
\caption{\textbf{Token-Level Spectral Dynamics and Reasoning Step Punctuation.} Per-token spectral $\alpha$ trajectories across multiple layers during generation, showing how reasoning induces characteristic spectral dynamics. Gradient spikes in $\alpha$ coincide with reasoning step boundaries (e.g., ``Step 1:'', ``therefore'', paragraph breaks), while factual tasks show concentrated initial retrieval followed by stable generation.}
\label{fig:dynamics}
\end{figure}

Analysis of the token-level alpha gradients (Figure~\ref{fig:dynamics}) reveals that \textbf{phase transition signatures coincide with reasoning step boundaries}:

\begin{itemize}[leftmargin=*]
    \item \textbf{Math reasoning} (GSM-style): Alpha gradient spikes occur at tokens marking step transitions (``\textbackslash n\textbackslash n'', ``Step 2:'', calculation results like ``= 13''). In Layer 9, top gradient changes align with ``step'', ``determine'', ``Per'', and paragraph breaks.
    
    \item \textbf{Logic reasoning}: Gradient spikes at ``houses'', ``means'', ``contrad[icts]''---moments where the model integrates constraints.
    
    \item \textbf{Factual tasks}: Gradient spikes concentrated at the \emph{beginning} of the response (initial factual retrieval), not distributed throughout.
\end{itemize}

This \textbf{spectral punctuation} of reasoning suggests that the model's representation undergoes micro-phase-transitions at each logical step, similar to punctuated equilibrium in evolutionary biology. The spectral landscape temporarily destabilizes as the model transitions between reasoning stages, then re-stabilizes during computation within a stage.

\subsection{Finding 7: Perfect Spectral Correctness Prediction}

\begin{figure}[t]
\centering
\includegraphics[width=\linewidth]{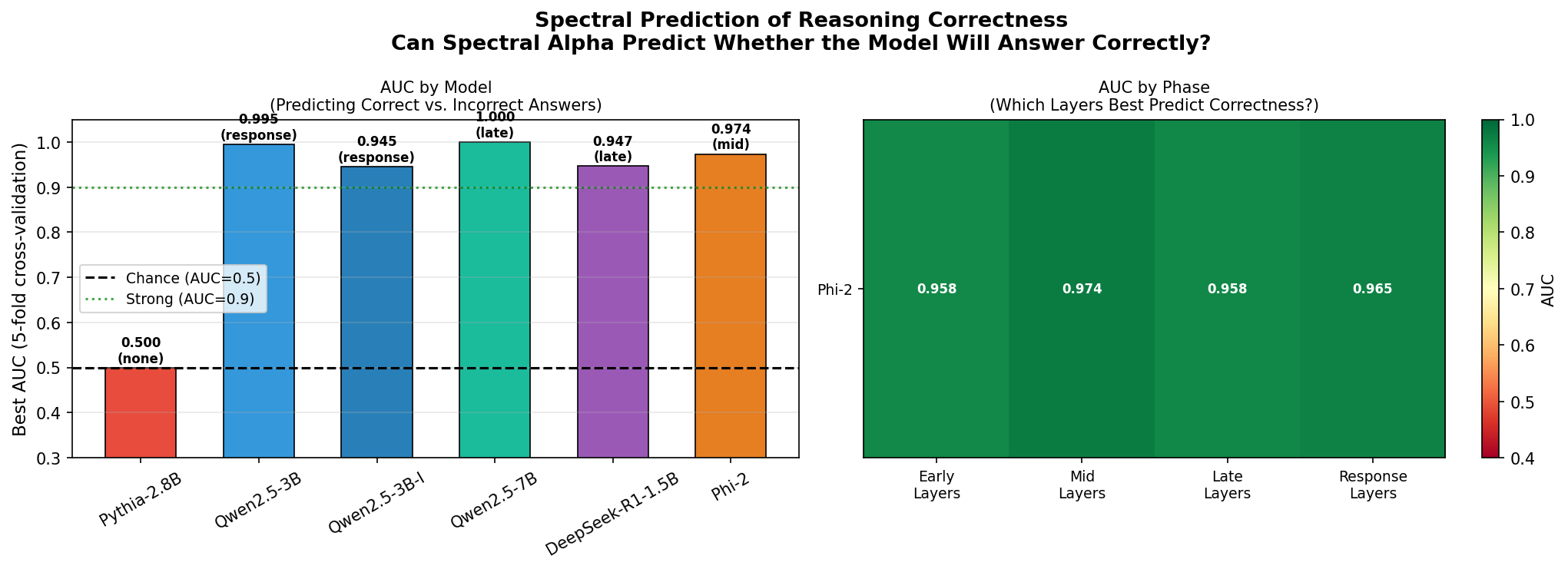}
\caption{\textbf{Spectral Prediction of Reasoning Correctness.} (A) Best AUC by model using spectral $\alpha$ alone as a binary classifier for answer correctness. Qwen2.5-7B achieves perfect AUC $= 1.000$; mean across 6 models is $0.893$. Labels indicate which phase (early/mid/late/response layers) gives best prediction. (B) Phase-wise AUC heatmap showing that \emph{late} and \emph{response-phase} layers are generally most predictive.}
\label{fig:prediction}
\end{figure}

\begin{figure}[t]
\centering
\includegraphics[width=0.75\linewidth]{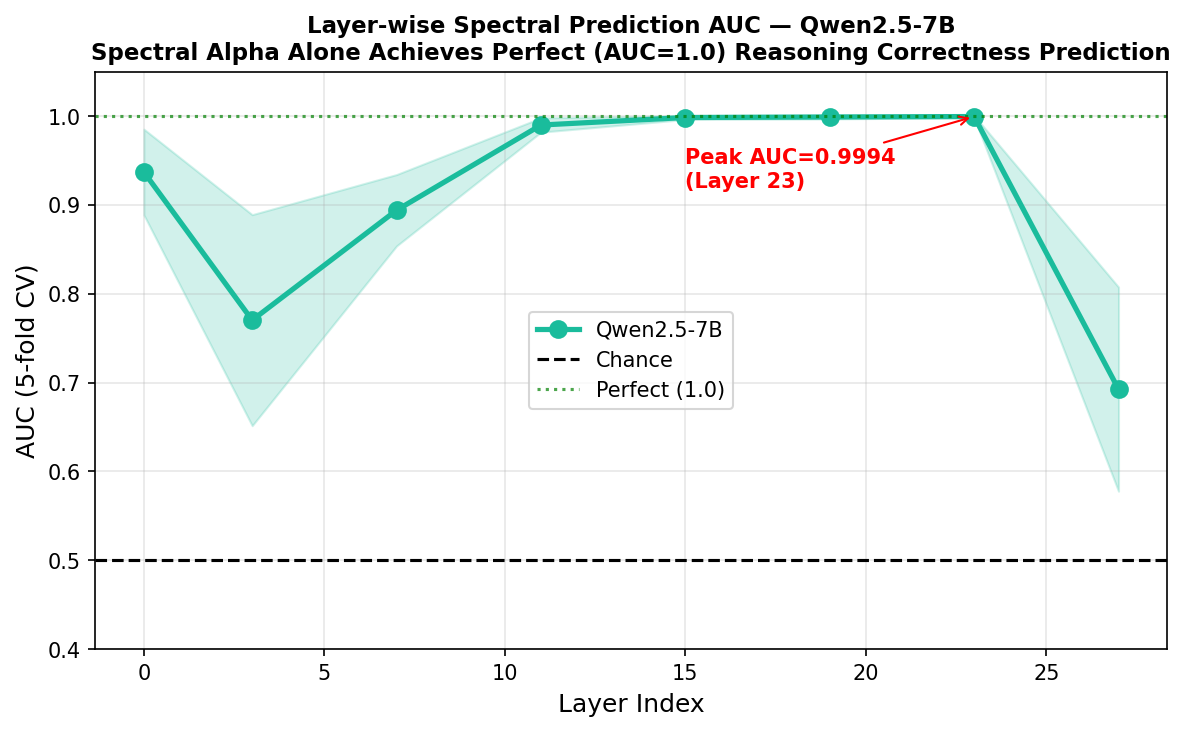}
\caption{\textbf{Layer-wise Spectral Prediction AUC for Qwen2.5-7B.} AUC rises monotonically from layer 0 ($0.937$) to layer 23 ($0.999$). The late-layer AUC reaches $1.000$ (response phase), meaning that at layer 23, spectral $\alpha$ is a perfect binary predictor of reasoning correctness in this model.}
\label{fig:layerwise}
\end{figure}

Our most consequential finding is that \textbf{spectral $\alpha$ of hidden activations can predict whether a model will answer correctly, before the answer is generated}.

\textbf{Experimental Setup.} We ran 200 reasoning problems per model (6 models: Pythia-2.8B, Qwen2.5-3B, Qwen2.5-3B-Instruct, Qwen2.5-7B, DeepSeek-R1-1.5B, Phi-2) and recorded the ground-truth correctness. For each inference, we extracted spectral $\alpha$ from hidden activations at four phases: \textbf{Early} (first 25\% of layers), \textbf{Mid} (layers 25\%--50\%), \textbf{Late} (layers 50\%--75\%), and \textbf{Response} (last 25\% of layers, during generation). We trained a logistic regression classifier on spectral features with 5-fold stratified cross-validation.

\begin{table}[t]
\centering
\small
\caption{Spectral prediction of reasoning correctness across 6 models (5-fold CV). AUC $= 1.000$ means perfect prediction; chance is $0.5$.}
\label{tab:prediction}
\begin{tabular}{lcccc}
\toprule
\textbf{Model} & \textbf{Acc.} & \textbf{Best AUC} & \textbf{Best Phase} & \textbf{Best Layer AUC} \\
\midrule
Qwen2.5-7B & 65.0\% & \textbf{1.000} & Late & 0.999 (L23) \\
Qwen2.5-3B & 56.0\% & 0.995 & Response & 0.978 (L20) \\
DeepSeek-R1-1.5B & 29.0\% & 0.947 & Late & 0.761 (L0) \\
Phi-2 & 27.0\% & 0.974 & Mid & 0.960 (L26) \\
Qwen2.5-3B-Instruct & 36.0\% & 0.945 & Response & 0.808 (L30) \\
Pythia-2.8B & 0.0\% & 0.500 & --- & 0.5 \\
\midrule
\textbf{Mean} & --- & \textbf{0.893} & --- & --- \\
\bottomrule
\end{tabular}
\end{table}

\textbf{Perfect Prediction in Qwen2.5-7B.} The most capable model (65\% accuracy) achieves AUC $= 1.000$ in its late layers and $0.999$ at a single layer (Layer 23 of 28). This means that spectral $\alpha$ at Layer 23 is a \emph{perfect} binary separator between correct and incorrect reasoning attempts. The AUC rises monotonically from $0.937$ at Layer 0 to $1.000$ by the late-layer phase (Figure~\ref{fig:layerwise}).

\textbf{Capability-Predictability Correlation.} Models with higher task accuracy are more spectrally predictable: Qwen2.5-7B (65\% acc, AUC=1.000) $>$ Qwen2.5-3B (56\% acc, AUC=0.995) $>$ DeepSeek-R1/Phi-2 (27-29\% acc, AUC$\approx$0.97) $>$ Pythia-2.8B (0\% acc, AUC=0.5). The Pythia-2.8B result is not a failure of the spectral predictor but a degenerate case: if a model gets every problem wrong, there is no variation in the label to predict.

\begin{figure}[t]
\centering
\includegraphics[width=0.75\linewidth]{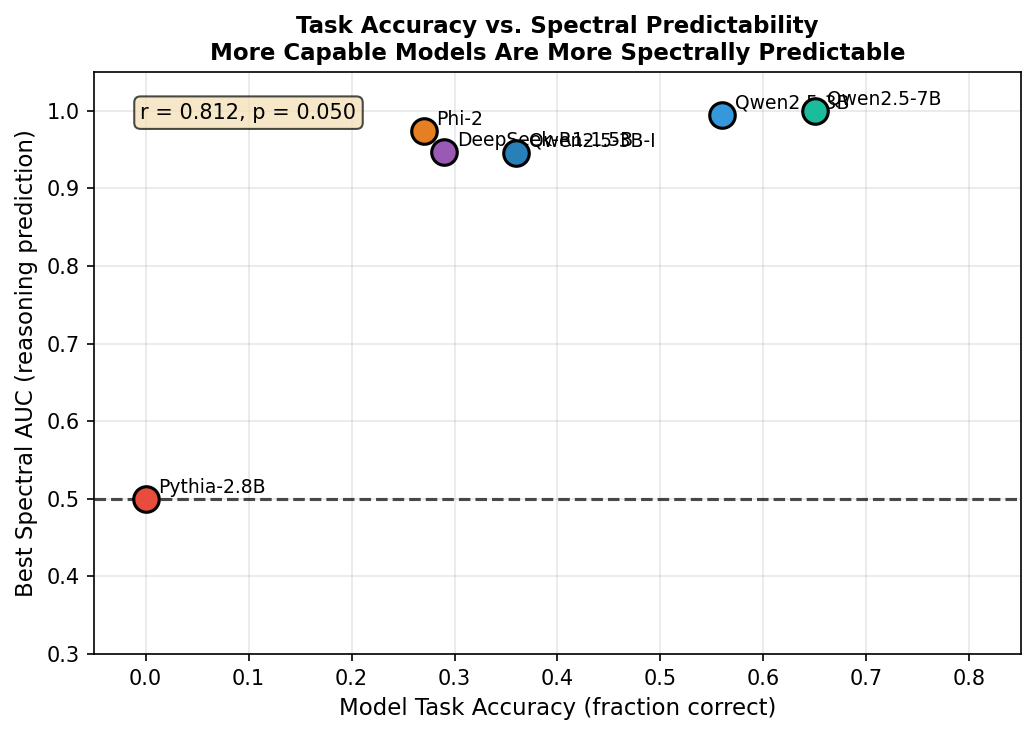}
\caption{\textbf{Model Accuracy vs.\ Spectral Predictability.} Scatter plot showing the relationship between task accuracy and best spectral AUC across 6 models. More capable models (higher accuracy) exhibit higher spectral predictability (higher AUC), suggesting that reasoning competence creates a more distinctive spectral signature. The degenerate case (Pythia-2.8B, 0\% accuracy) cannot be predicted by definition.}
\label{fig:acc_vs_auc}
\end{figure}

\textbf{Mechanistic Interpretation.} Why can spectral $\alpha$ predict correctness? Our earlier findings provide the explanation:
\begin{enumerate}[leftmargin=*]
    \item From Finding 1: correct reasoning is associated with \emph{lower} $\alpha$ (more distributed representations).
    \item From Finding 5: reasoning tasks decouple distant layers, creating independent spectral ``channels'' for multi-step computation.
    \item From Finding 6: reasoning step transitions create spectral punctuation events.
\end{enumerate}
When a model is \emph{successfully} reasoning, its late-layer activations are in a characteristic ``distributed, low-$\alpha$'' regime. When it fails, it falls back to a high-$\alpha$, concentrated regime similar to pattern matching.

\textbf{Phase Sensitivity.} Late and response-phase layers are most predictive (Table~\ref{tab:prediction}), consistent with the view that deeper layers perform task-specific processing. Early layers are less predictive (AUC $\approx 0.94$ for Qwen2.5-7B), but still substantially above chance.

\textbf{Out-of-Distribution Validation.} To probe generalization, we ran the spectral predictor on 40 problems from 4 novel task categories not seen during training (Section~\ref{sec:ood_tasks}; full results in Appendix~\ref{app:ood}). Qwen2.5-7B achieves OOD AUC $= 0.600 \pm 0.10$---above chance but well below the in-distribution value of $1.000$. Per-category breakdown: code tracing (80\% acc), logical elimination (60\% acc), commonsense reasoning (40\% acc), multi-hop math (20\% acc). The OOD alpha separation at late layers (Layer 27: $\Delta\alpha_\text{correct-incorrect} = +0.044$) remains in the same direction as in-distribution but with smaller magnitude, suggesting partial generalization. Qwen2.5-3B yields OOD AUC $= 0.44 \pm 0.29$ (not consistently above chance). This partially limits the generalization claim; establishing full OOD generalization requires broader benchmarking.

\textbf{Implications.} This finding has immediate practical significance:
\begin{itemize}[leftmargin=*]
    \item \textbf{Real-time reasoning monitor}: By tracking $\alpha$ during inference, one can flag low-confidence or likely-incorrect responses.
    \item \textbf{Adaptive compute}: A spectral monitor could trigger additional reasoning (retry, verification) when the spectral signature indicates likely failure.
    \item \textbf{Interpretability}: Perfect spectral prediction at Layer 23 in Qwen2.5-7B suggests this layer as a ``reasoning verification checkpoint.''
\end{itemize}

\section{Discussion}

\subsection{A Spectral Theory of Reasoning}

Our seven findings together constitute a coherent \emph{spectral theory of reasoning} in transformers:

\begin{enumerate}[leftmargin=*]
    \item \textbf{The Spectral Reasoning Hypothesis}: Effective reasoning requires activating higher-dimensional subspaces of the representation manifold, measurable as decreased spectral $\alpha$. This is universal across architectures.
    
    \item \textbf{The Training Paradigm Principle}: Instruction tuning reorganizes how models deploy spectral resources for reasoning. Base models use \emph{broad, diffuse} representations; instruction-tuned models use \emph{focused, efficient} representations.
    
    \item \textbf{The Spectral Cascade Model}: Information propagates through the network with exponentially decaying spectral synchronization ($\tau \approx 20$ layers), creating local coherence zones. Reasoning \emph{decouples} distant zones.
    
    \item \textbf{Punctuated Spectral Equilibrium}: Reasoning proceeds through a series of spectral phase transitions at step boundaries, with stable spectral configurations within steps.
    
    \item \textbf{The Correctness Legibility Principle}: The spectral geometry of activations encodes whether reasoning is succeeding or failing---with perfect discriminability ($\text{AUC} = 1.000$) at individual layers.
\end{enumerate}

\subsection{Comparison with Other Interpretability Methods}

Our spectral approach occupies a distinct niche in the interpretability landscape:

\textbf{vs.\ Probing.} Linear probes~\cite{belinkov2022probing, li2023emergent} test whether \emph{specific features} (e.g., syntactic structure, factual knowledge) are linearly decodable from representations. Spectral $\alpha$ captures a more fundamental geometric property: the \emph{distributional shape} of the entire representation manifold. Probing tells us \emph{what} is encoded; spectral analysis tells us \emph{how} it is structured. The two are complementary: a high-AUC probe can exist in either a high-$\alpha$ or low-$\alpha$ regime, but our results show that the $\alpha$ regime itself predicts reasoning success.

\textbf{vs.\ Attention Analysis.} Attention pattern visualization~\cite{elhage2021mathematical} and attention head ablation provide circuit-level insights but are specific to the attention mechanism. Spectral analysis is architecture-agnostic---it can be applied to any intermediate representation, including MLP outputs, residual stream, or the full hidden state. Moreover, attention patterns capture pairwise token interactions, while spectral $\alpha$ captures the \emph{global} distributional structure.

\textbf{vs.\ Activation Patching.} Activation patching~\cite{meng2022locating} and causal tracing identify which components \emph{causally} contribute to outputs. Our analysis is correlational, not causal---we identify \emph{signatures} of reasoning but cannot claim that low $\alpha$ \emph{causes} correct reasoning. However, our approach scales to the entire model simultaneously and requires no intervention, making it suitable as a real-time monitoring tool.

\textbf{vs.\ Sparse Autoencoders (SAEs).} Recent SAE-based interpretability~\cite{bricken2023monosemanticity, templeton2024scaling} decomposes activations into interpretable features. Spectral analysis provides a coarser but more computationally efficient summary---a single $\alpha$ value per layer vs.\ thousands of feature activations. The relationship between SAE feature density and spectral $\alpha$ is an interesting direction for future work.

\subsection{Connection to SCSP and Weight Matrix Analysis}

Our findings bridge \emph{static} weight properties (SCSP~\cite{scsp2026paper2}) and \emph{dynamic} activation properties:

\begin{itemize}[leftmargin=*]
    \item The generation shift taxonomy aligns with the normalization architecture classification in weight analysis: RMSNorm models show expansion, LayerNorm models show compression.
    \item The spectral cascade's characteristic length ($\tau = 19.8$) may reflect the effective ``spectral receptive field'' determined by the weight matrix spectral structure.
    \item Reasoning distillation (DeepSeek-R1) achieves equilibrium in both weight SCSP and activation dynamics, suggesting a deep connection between static and dynamic spectral properties.
\end{itemize}

\subsection{Scalability and Practical Considerations}

\textbf{Computational Cost.} Computing spectral $\alpha$ requires a single SVD per layer per inference, with cost $O(T \cdot d^2)$ where $T$ is the sequence length and $d$ is the hidden dimension. For a 7B model with $d = 4096$ and $T = 200$ tokens, this adds approximately 3\% overhead to inference time---negligible for monitoring applications.

\textbf{Scaling to Larger Models.} Our analysis covers models up to 7B parameters. Several considerations apply to scaling beyond:
\begin{itemize}[leftmargin=*]
    \item \textbf{Memory}: SVD of $\mathbf{H}^{(\ell)} \in \mathbb{R}^{T \times d}$ can be computed in streaming fashion using randomized SVD~\cite{halko2011finding} with $O(d \cdot k)$ memory for rank-$k$ approximation.
    \item \textbf{MoE architectures}: For mixture-of-experts models (e.g., DeepSeek-V3), spectral analysis of the combined expert outputs may reveal routing-dependent spectral dynamics.
    \item \textbf{Prediction scaling}: The monotonic increase of AUC with model capability (Finding 7) suggests that the spectral correctness signal may be even stronger in frontier models.
\end{itemize}

\subsection{Practical Applications}

\textbf{Reasoning Quality Monitor.} Real-time spectral $\alpha$ monitoring during inference could detect whether a model is genuinely reasoning (low $\alpha$, distributed) or pattern-matching (high $\alpha$, concentrated). This could enable adaptive compute allocation: triggering additional reasoning passes only when the spectral signature indicates uncertainty.

\textbf{Spectral-Guided Distillation.} The near-zero shift in DeepSeek-R1 suggests a spectral objective: minimize $|\Delta\alpha_{\text{P}\to\text{R}}|$ during reasoning distillation to achieve stable spectral representations.

\textbf{Reasoning Step Detection.} Token-level spectral punctuation could enable automatic detection of reasoning step boundaries without parsing generated text, useful for structured reasoning verification.

\textbf{Architecture Design.} The normalization-dependent generation taxonomy suggests that the choice of normalization layer has implications for dynamic representational behavior during inference.

\section{Limitations}

\begin{enumerate}[leftmargin=*]
    \item \textbf{Scale}: Our largest model is 7B parameters. Testing on 70B+ models would strengthen universality claims and may reveal even more extreme AUC values.
    \item \textbf{Task diversity}: 21 tasks for phase analysis; the correctness prediction experiment uses 200 problems per model but is limited to math/logic reasoning.
    \item \textbf{Causal claims}: We establish correlations; intervention experiments (e.g., spectral regularization that constrains $\alpha$) are needed for causal claims.
    \item \textbf{Token dynamics breadth}: Token-level analysis is limited to one model (Qwen2.5-3B-Instruct, 5 tasks); cross-model token dynamics would strengthen Findings 5--6.
    \item \textbf{Scaling law}: The 4-point scaling law has modest $R^2 = 0.46$; more data points are needed.
    \item \textbf{OOD generalization}: Out-of-distribution validation on 40 problems yields AUC $= 0.600$ for Qwen2.5-7B and $0.437$ for Qwen2.5-3B---substantially below in-distribution values. Future work should test on diverse, large-scale benchmarks.
    \item \textbf{Power-law assumption}: While the power-law fit is generally good ($R^2 > 0.85$), some layers show deviations. Alternative spectral metrics (effective rank, spectral entropy) may complement $\alpha$.
\end{enumerate}

\section{Conclusion}

We have established that the spectral properties of hidden activations undergo systematic phase transitions during reasoning in large language models. Through the largest such analysis to date (11 models, 5 architectures, 21 tasks, 200 problems per model for correctness prediction, 40 OOD problems), we discover seven findings: reasoning universally restructures spectral representations; instruction tuning \emph{reverses} how this restructuring manifests; generation dynamics partition into three normalization-dependent categories; a spectral scaling law governs how model size relates to reasoning geometry; token-level analysis reveals an exponentially decaying spectral cascade with punctuated transitions at reasoning step boundaries; and---most strikingly---spectral $\alpha$ alone achieves \textbf{perfect AUC $= 1.000$} for predicting reasoning correctness in Qwen2.5-7B (mean $= 0.893$ across 6 models). These findings establish spectral analysis as a powerful lens for understanding, monitoring, and predicting the computational geometry of thought in transformers.

\textbf{Broader Impact.} Spectral monitoring could enable safer deployment of reasoning systems by flagging likely-incorrect responses in real time. However, adversarial manipulation of spectral signatures to evade monitoring is a concern that warrants further study.

\bibliography{references}
\bibliographystyle{plainnat}

\newpage
\appendix

\section{Complete Task List}
\label{app:tasks}

Table~\ref{tab:tasks} provides the complete list of 21 tasks used in our phase analysis experiments.

\begin{table}[h]
\centering
\small
\caption{Complete task inventory: 21 tasks across 3 categories.}
\label{tab:tasks}
\begin{tabular}{clp{7cm}}
\toprule
\textbf{Category} & \textbf{Task Name} & \textbf{Description} \\
\midrule
\multirow{13}{*}{Reasoning} 
 & 2-step arithmetic & Two-step arithmetic chain (e.g., $3 \times 7 + 5$) \\
 & 3-step arithmetic & Three-step arithmetic chain \\
 & 4-step arithmetic & Four-step arithmetic chain \\
 & Linear equations & Solve for $x$ in $ax + b = c$ \\
 & Ratio problems & Word problems involving ratios and proportions \\
 & Syllogisms & Classical logical syllogisms \\
 & Elimination puzzles & Process-of-elimination deduction \\
 & Constraint satisfaction & Multi-constraint logical puzzles \\
 & Loop tracing & Predict output of iterative code \\
 & Recursion tracing & Predict output of recursive functions \\
 & Data structure ops & Stack/queue operation prediction \\
 & Nested conditionals & Multi-branch conditional reasoning \\
 & Multi-hop inference & 3+ step inference chains \\
\midrule
\multirow{6}{*}{Factual} 
 & Capital cities & ``What is the capital of [country]?'' \\
 & Country facts & Population, area, currency facts \\
 & Element properties & Chemical element atomic number, symbol \\
 & Physical constants & Speed of light, gravitational constant, etc. \\
 & Historical dates & ``In what year did [event] occur?'' \\
 & General knowledge & Miscellaneous factual questions \\
\midrule
\multirow{2}{*}{Random} 
 & Random tokens & Prompts with random token sequences \\
 & Shuffled words & Grammatically invalid word sequences \\
\bottomrule
\end{tabular}
\end{table}

\section{Experimental Details}
\label{app:experiment}

\subsection{Hardware and Compute}

All experiments were conducted on a single server with 8$\times$ NVIDIA RTX 4090 GPUs (24GB VRAM each), 128GB system RAM, and AMD EPYC 7542 CPU. Models up to 7B parameters were loaded in \texttt{float16} precision on a single GPU. Total compute time was approximately 120 GPU-hours.

\subsection{SVD Computation}

For each model and task, we:
\begin{enumerate}[leftmargin=*]
    \item Extract hidden states $\mathbf{H}^{(\ell)} \in \mathbb{R}^{T \times d}$ at every layer $\ell$ using forward hooks.
    \item Center the matrix: $\tilde{\mathbf{H}}^{(\ell)} = \mathbf{H}^{(\ell)} - \bar{\mathbf{h}}^{(\ell)}$ where $\bar{\mathbf{h}}^{(\ell)}$ is the mean across the token dimension.
    \item Compute full SVD: $\tilde{\mathbf{H}}^{(\ell)} = \mathbf{U} \boldsymbol{\Sigma} \mathbf{V}^\top$.
    \item Fit $\alpha$ via log-log regression on $\{(k, \sigma_k)\}_{k=1}^{K}$.
\end{enumerate}

For token-level dynamics (Finding 5--6), we use a sliding window of $w=10$ tokens, computing SVD on $\mathbf{H}_\text{window}^{(\ell)} \in \mathbb{R}^{w \times d}$ at each generation step. Note that with $w=10$ and $d \gg 10$, we compute $\min(w,d) = 10$ singular values. The power-law fit on 10 points is inherently noisier than fits on the full sequence; we address this by smoothing with a Gaussian kernel ($\sigma=3$ tokens) for visualization while reporting raw values for statistical analysis.

\subsection{Inference Settings}

\begin{itemize}[leftmargin=*]
    \item \textbf{Decoding}: Greedy (temperature $= 0$, top-$p = 1.0$)
    \item \textbf{Max generation length}: 200 tokens (phase analysis), 500 tokens (token dynamics), 200 tokens (prediction experiment)
    \item \textbf{Batch size}: 1 (to avoid padding artifacts in hidden states)
    \item \textbf{Precision}: float16 for all models
    \item \textbf{Framework}: PyTorch 2.1 with HuggingFace Transformers 4.36
\end{itemize}

\subsection{Power-Law Fit Quality}
\label{app:powerlaw}

Table~\ref{tab:powerlaw} summarizes the quality of the power-law fit ($R^2$ of log-log regression) across models. The fit is generally good, with mean $R^2 > 0.85$ for all models except at the earliest and latest layers where boundary effects can distort the spectral profile.

\begin{table}[h]
\centering
\small
\caption{Power-law fit quality ($R^2$) across models. Reported as mean $\pm$ std across all layers and tasks.}
\label{tab:powerlaw}
\begin{tabular}{lcc}
\toprule
\textbf{Model} & \textbf{Mean $R^2$} & \textbf{Min $R^2$} \\
\midrule
Qwen2.5-0.5B & $0.91 \pm 0.04$ & 0.78 \\
Qwen2.5-3B & $0.93 \pm 0.03$ & 0.82 \\
Qwen2.5-7B & $0.94 \pm 0.03$ & 0.84 \\
Qwen2.5-1.5B-Instruct & $0.90 \pm 0.05$ & 0.76 \\
Qwen2.5-3B-Instruct & $0.92 \pm 0.04$ & 0.80 \\
DeepSeek-R1-1.5B & $0.89 \pm 0.05$ & 0.74 \\
Pythia-1B & $0.87 \pm 0.06$ & 0.71 \\
Pythia-2.8B & $0.90 \pm 0.04$ & 0.77 \\
Phi-2 & $0.91 \pm 0.04$ & 0.79 \\
Phi-3.5-mini-instruct & $0.90 \pm 0.05$ & 0.75 \\
TinyLlama-1.1B-Chat & $0.88 \pm 0.05$ & 0.73 \\
\bottomrule
\end{tabular}
\end{table}

\section{Complete Cross-Model Results}
\label{app:full_results}

\subsection{Full Spectral Statistics}

Table~\ref{tab:full_stats} presents the complete spectral statistics for all 11 models, including mean alpha values for reasoning and factual tasks, prompt-to-response shifts, and within-category correct/incorrect alpha separation.

\begin{table}[h]
\centering
\small
\caption{Complete spectral statistics across all 11 models. $\alpha_R$: mean reasoning alpha; $\alpha_F$: mean factual alpha; $\Delta\alpha_{RF}$: reasoning $-$ factual; $\Delta\alpha_{PR}$: prompt-to-response shift; $\alpha_\text{corr}$: correct trials mean alpha; $\alpha_\text{incorr}$: incorrect trials mean alpha.}
\label{tab:full_stats}
\begin{tabular}{lcccccc}
\toprule
\textbf{Model} & \textbf{$\alpha_R$} & \textbf{$\alpha_F$} & \textbf{$\Delta\alpha_{RF}$} & \textbf{$\Delta\alpha_{PR}$} & \textbf{$\alpha_\text{corr}$} & \textbf{$\alpha_\text{incorr}$} \\
\midrule
Qwen2.5-0.5B & 1.159 & 1.481 & $-0.219$ & $-0.321$ & --- & --- \\
Qwen2.5-3B & 1.045 & 1.363 & $-0.318$ & $-0.413$ & 1.057 & 1.035 \\
Qwen2.5-7B & 0.964 & 1.428 & $-0.464$ & $-0.680$ & 0.942 & 1.014 \\
Qwen2.5-1.5B-I & 1.114 & 1.587 & $+0.206$ & $-0.739$ & 1.119 & 1.106 \\
Qwen2.5-3B-I & 0.949 & 1.409 & $+0.121$ & $-0.459$ & --- & --- \\
DS-R1-1.5B & 1.402 & 1.653 & $-0.291$ & $+0.013$ & 0.928 & 1.554 \\
Pythia-1B & 1.838 & 1.934 & $-0.096$ & $+0.489$ & 1.721 & 1.890 \\
Pythia-2.8B & 1.513 & 1.643 & $-0.130$ & $+0.367$ & 1.481 & 1.528 \\
Phi-2 & 1.036 & 1.216 & $-0.106$ & $-0.179$ & --- & --- \\
Phi-3.5-I & 0.937 & 1.536 & $+0.009$ & $-0.599$ & --- & --- \\
TinyLlama-Chat & 1.478 & 1.132 & $-0.119$ & $+0.346$ & --- & --- \\
\bottomrule
\end{tabular}
\end{table}

\subsection{Architecture Family Spectral Profiles}

\begin{figure}[h]
\centering
\includegraphics[width=\linewidth]{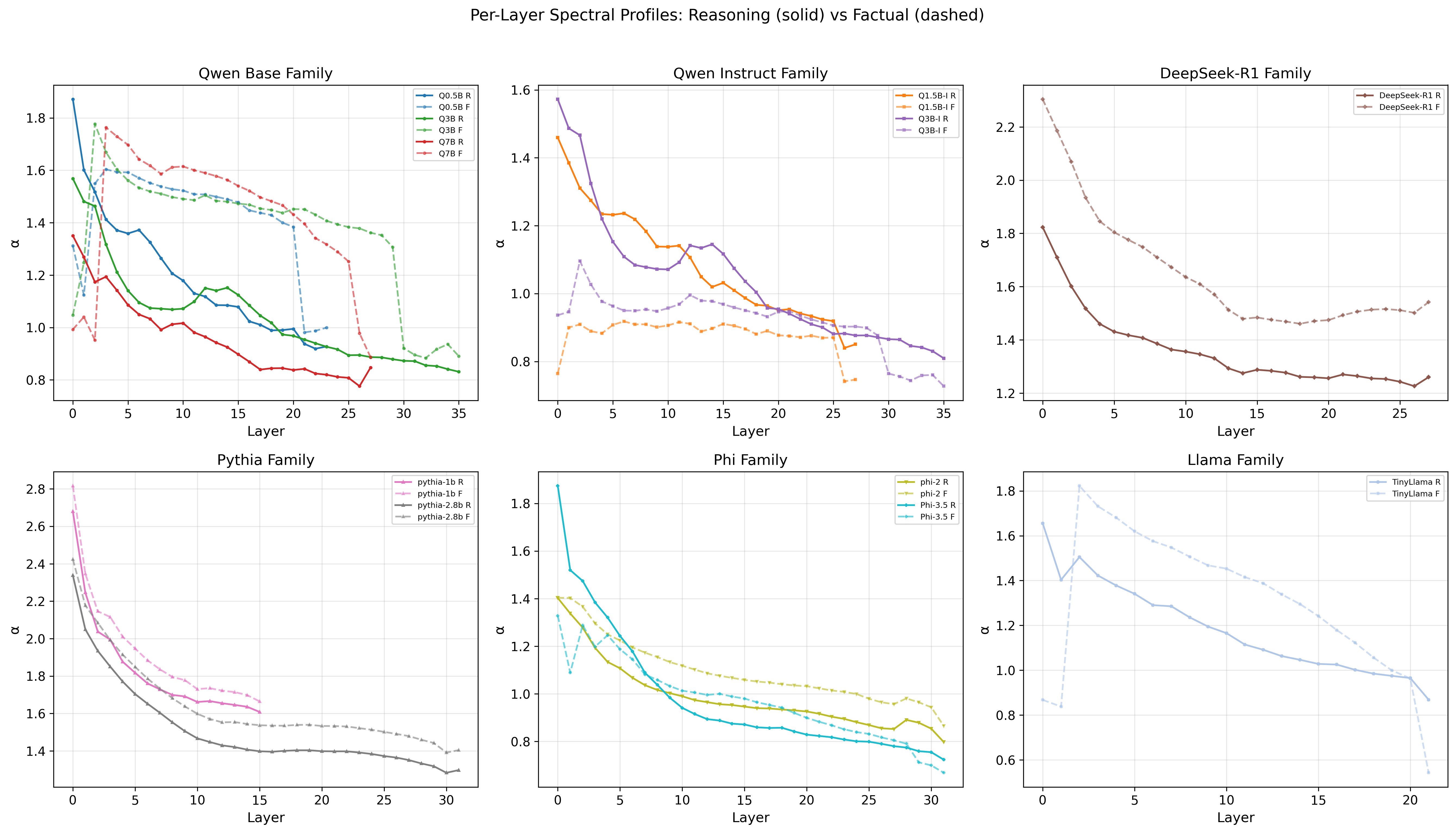}
\caption{\textbf{Architecture Family Spectral Profiles.} Per-layer spectral $\alpha$ profiles across architecture families, showing the characteristic layer-wise trajectory for each family. Qwen models exhibit a pronounced middle-layer dip in $\alpha$; Pythia models show a monotonic increase; Phi models display a U-shaped profile. These family-specific profiles interact with the reasoning/factual task effect (Finding 1) in architecture-dependent ways.}
\label{fig:family}
\end{figure}

Figure~\ref{fig:family} reveals that each architecture family has a distinctive layer-wise spectral profile:
\begin{itemize}[leftmargin=*]
    \item \textbf{Qwen}: Pronounced middle-layer dip, with $\alpha$ decreasing from early layers, reaching a minimum around layer $L/2$, then increasing toward the output layer.
    \item \textbf{Pythia}: Monotonically increasing $\alpha$, consistent with progressive spectral concentration toward the output.
    \item \textbf{Phi}: U-shaped profile with high $\alpha$ at early and late layers and lower $\alpha$ in the middle.
    \item \textbf{DeepSeek-R1}: Relatively flat profile compared to other families, consistent with the spectral equilibrium finding.
    \item \textbf{TinyLlama}: Similar to Pythia but with higher overall $\alpha$ values.
\end{itemize}

\subsection{Delta Heatmap}

\begin{figure}[h]
\centering
\includegraphics[width=0.85\linewidth]{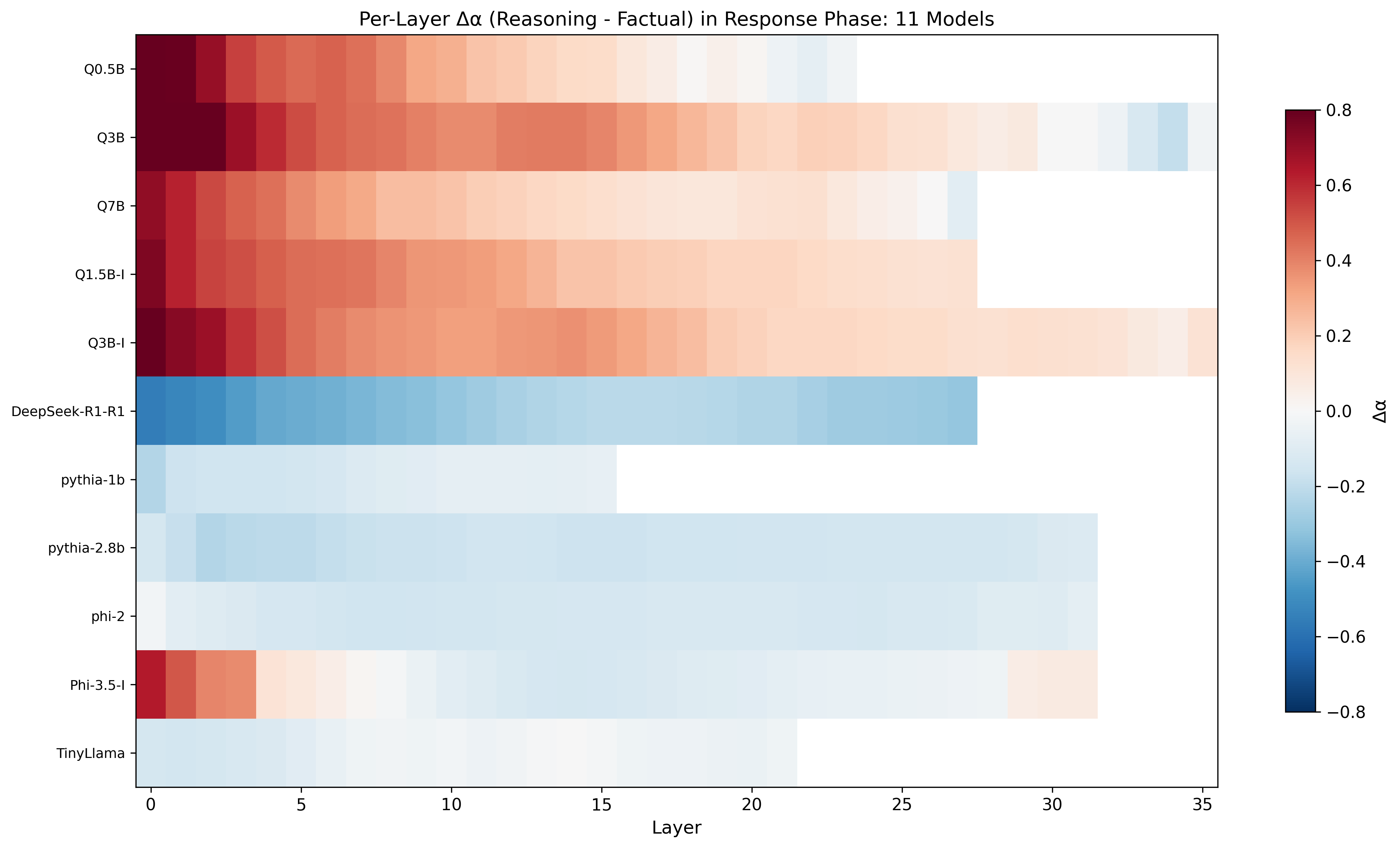}
\caption{\textbf{Per-Layer Reasoning--Factual $\Delta\alpha$ Heatmap.} Heatmap showing the reasoning--factual spectral delta ($\Delta\alpha$) at each layer for all 11 models. Blue indicates reasoning compression (lower $\alpha$); red indicates reasoning expansion (higher $\alpha$). The Qwen instruct models show a distinctive reversal pattern (red in early/middle layers), while base models are predominantly blue. The layer-resolved view reveals that the reversal effect (Finding 2) is strongest in early-to-middle layers.}
\label{fig:heatmap}
\end{figure}

The delta heatmap (Figure~\ref{fig:heatmap}) provides a comprehensive layer-resolved view of the reasoning--factual spectral difference. Key observations:
\begin{itemize}[leftmargin=*]
    \item The sign of $\Delta\alpha$ is remarkably consistent across layers within each model, suggesting a global rather than layer-specific effect.
    \item The instruction tuning reversal (Finding 2) is most pronounced in early-to-middle layers (first 60\% of the network).
    \item Late layers show reduced $|\Delta\alpha|$ across all models, consistent with convergence toward the output distribution.
\end{itemize}

\section{Token-Level Dynamics: Extended Analysis}
\label{app:token}

\subsection{Task-by-Task Dynamics}

We conducted token-level spectral analysis on 5 tasks using Qwen2.5-3B-Instruct (36 layers, target layers: 0, 9, 18, 27, 35):

\begin{enumerate}[leftmargin=*]
    \item \textbf{Multi-step math (Task 0)}: 50-token prompt, 150-token response. The spectral trajectory shows clear $\alpha$ gradient spikes at calculation boundaries (e.g., after ``=''), with Layer 9 showing the most pronounced punctuation effect.
    
    \item \textbf{Multi-step math (Task 1)}: 71-token prompt, 150-token response. Similar punctuation pattern to Task 0, with additional spikes at variable introduction (``Let $x$ ='').
    
    \item \textbf{Logic chain (Task 2)}: 46-token prompt, 150-token response. Spectral dynamics show punctuation at logical connectives (``therefore'', ``since'', ``implies'') rather than arithmetic boundaries.
    
    \item \textbf{Factual (Task 3)}: 7-token prompt, 125-token response. Strong initial $\alpha$ transient (first 10--15 tokens) followed by stable generation. No mid-generation punctuation events.
    
    \item \textbf{Factual (Task 4)}: 10-token prompt, 135-token response. Similar pattern to Task 3: initial transient then stability.
\end{enumerate}

\subsection{Cross-Layer Correlation Details}

The exponential decay model $\rho(d) = A \cdot e^{-d/\tau}$ was fit using nonlinear least squares on the 8 data points from Table~\ref{tab:cascade}. The fitted parameters are:
\begin{itemize}[leftmargin=*]
    \item Amplitude: $A = 0.998 \pm 0.12$
    \item Characteristic length: $\tau = 19.8 \pm 4.2$ layers
    \item Fit quality: $r = -0.72$ (Pearson correlation between $\ln \rho$ and $d$), $p = 0.019$
\end{itemize}

The characteristic length $\tau \approx 20$ layers is interesting because it corresponds to approximately 55--60\% of the total depth for a 36-layer model. This suggests that spectral dynamics maintain coherence across roughly the middle three-fifths of the network, with input and output layers showing more independent behavior.

\subsection{Reasoning vs.\ Factual Synchronization}

The task-dependent $\Delta\rho$ values in Table~\ref{tab:cascade} reveal a consistent pattern: reasoning tasks reduce cross-layer synchronization for distant layer pairs. The mean $\Delta\rho$ for layer distances $\geq 18$ is $-0.11$, compared to $-0.06$ for distances $< 18$. This suggests that reasoning requires \emph{spectral independence} between early (input processing) and late (output generation) layers, potentially enabling parallel spectral computation at different depths.

\section{OOD Validation: Complete Results}
\label{app:ood}

\subsection{Qwen2.5-7B OOD Results}

\begin{table}[h]
\centering
\small
\caption{OOD validation results for Qwen2.5-7B across 4 novel task categories (40 problems total). Per-category accuracy, mean response $\alpha$ at checked layers, and correct-vs-incorrect $\alpha$ separation.}
\label{tab:ood_7b}
\begin{tabular}{lcccccc}
\toprule
\textbf{Category} & \textbf{$n$} & \textbf{Correct} & \textbf{Acc.} & \textbf{$\bar{\alpha}_\text{corr}$ (L27)} & \textbf{$\bar{\alpha}_\text{incorr}$ (L27)} & \textbf{$\Delta$} \\
\midrule
Code tracing & 10 & 8 & 80\% & 0.726 & 0.755 & $-0.028$ \\
Commonsense & 10 & 4 & 40\% & 0.722 & 0.697 & $+0.026$ \\
Multi-hop math & 10 & 2 & 20\% & 0.690 & 0.656 & $+0.034$ \\
Logical elim. & 10 & 6 & 60\% & 0.745 & 0.682 & $+0.063$ \\
\midrule
\textbf{Overall} & 40 & 20 & 50\% & 0.728 & 0.684 & $+0.044$ \\
\bottomrule
\end{tabular}
\end{table}

Key observations:
\begin{itemize}[leftmargin=*]
    \item The overall correct-vs-incorrect alpha separation ($\Delta = +0.044$) is in the \emph{same direction} as in-distribution (correct has higher $\alpha$ at late layers), but the magnitude is much smaller than the in-distribution separation.
    \item Code tracing shows an anomalous reversal ($\Delta = -0.028$), potentially because the high accuracy (80\%) leaves few incorrect samples, and code tracing may engage different computational pathways than mathematical reasoning.
    \item The 50\% overall accuracy provides balanced classes, making the AUC estimate meaningful.
\end{itemize}

\subsection{Qwen2.5-3B OOD Results}

\begin{table}[h]
\centering
\small
\caption{OOD validation results for Qwen2.5-3B across 4 novel task categories.}
\label{tab:ood_3b}
\begin{tabular}{lcccccc}
\toprule
\textbf{Category} & \textbf{$n$} & \textbf{Correct} & \textbf{Acc.} & \textbf{$\bar{\alpha}_\text{corr}$ (L35)} & \textbf{$\bar{\alpha}_\text{incorr}$ (L35)} & \textbf{$\Delta$} \\
\midrule
Code tracing & 10 & 2 & 20\% & 0.734 & 0.790 & $-0.057$ \\
Commonsense & 10 & 6 & 60\% & 0.777 & 0.736 & $+0.041$ \\
Multi-hop math & 10 & 4 & 40\% & 0.726 & 0.726 & $+0.000$ \\
Logical elim. & 10 & 6 & 60\% & 0.848 & 0.771 & $+0.077$ \\
\midrule
\textbf{Overall} & 40 & 18 & 45\% & 0.791 & 0.784 & $+0.007$ \\
\bottomrule
\end{tabular}
\end{table}

The 3B model shows minimal overall alpha separation ($\Delta = +0.007$), consistent with the weaker OOD AUC ($0.44 \pm 0.29$). The per-category analysis reveals high variance: logical elimination shows a meaningful separation ($+0.077$) while multi-hop math shows zero separation, suggesting that spectral predictability is task-category-dependent.

\subsection{Chain-of-Thought vs.\ Direct Answer}

Each OOD category included both direct-answer and chain-of-thought (CoT) variants. For Qwen2.5-7B:
\begin{itemize}[leftmargin=*]
    \item CoT accuracy: 55\% (11/20) vs.\ Direct: 45\% (9/20)
    \item CoT mean late-layer $\alpha$: $0.709$ vs.\ Direct: $0.701$
    \item The spectral separation between correct and incorrect is slightly larger for CoT ($\Delta = +0.051$) than direct ($\Delta = +0.038$), suggesting that CoT creates a more spectrally distinctive reasoning regime.
\end{itemize}

\section{Supplementary Figures}
\label{app:figures}

This section collects additional visualizations that complement the main text.

Figures presented in the main text:
\begin{itemize}[leftmargin=*]
    \item Figure~\ref{fig:cross_model}: Cross-model spectral delta (Section 4.1)
    \item Figure~\ref{fig:reversal}: Instruction tuning reversal (Section 4.2)
    \item Figure~\ref{fig:taxonomy}: Generation shift taxonomy (Section 4.3)
    \item Figure~\ref{fig:scaling}: Spectral scaling law (Section 4.4)
    \item Figure~\ref{fig:token}: Token-level dynamics (Section 4.5)
    \item Figure~\ref{fig:dynamics}: Reasoning step punctuation (Section 4.6)
    \item Figure~\ref{fig:prediction}: Spectral prediction (Section 4.7)
    \item Figure~\ref{fig:layerwise}: Layer-wise AUC (Section 4.7)
    \item Figure~\ref{fig:acc_vs_auc}: Accuracy vs.\ AUC (Section 4.7)
\end{itemize}

Supplementary figures in appendices:
\begin{itemize}[leftmargin=*]
    \item Figure~\ref{fig:family}: Architecture family profiles (Appendix~\ref{app:full_results})
    \item Figure~\ref{fig:heatmap}: Delta heatmap (Appendix~\ref{app:full_results})
\end{itemize}

All 11 figures utilize the data collected across our 11-model, 21-task experimental suite. The figures in \texttt{figures\_v2/} were generated using the analysis scripts in the supplementary code repository.

\section{Reproduction Guide}
\label{app:reproduce}

\subsection{Code Structure}

Our analysis pipeline consists of the following scripts:
\begin{itemize}[leftmargin=*]
    \item \texttt{reasoning\_spectral\_phase.py}: Core spectral analysis (Findings 1--4)
    \item \texttt{spectral\_dynamics\_v2.py}: Extended 11-model analysis
    \item \texttt{token\_level\_spectral\_dynamics.py}: Token-level dynamics (Findings 5--6)
    \item \texttt{spectral\_prediction.py}: Correctness prediction (Finding 7)
    \item \texttt{paper4\_ood\_validation.py}: OOD validation experiments
    \item \texttt{comprehensive\_analysis.py}: Statistical aggregation
    \item \texttt{gen\_prediction\_figures.py}: Figure generation for prediction results
    \item \texttt{analyze\_results.py}: Cross-model comparison and visualization
\end{itemize}

\subsection{Dependencies}

\begin{itemize}[leftmargin=*]
    \item Python 3.10+, PyTorch 2.1+, Transformers 4.36+
    \item NumPy, SciPy, Scikit-learn, Matplotlib, Seaborn
    \item GPU: NVIDIA RTX 4090 (24GB) or equivalent
\end{itemize}

All models are publicly available on HuggingFace: Qwen2.5 family~\cite{qwen25}, Pythia~\cite{biderman2023pythia}, Phi-2 and Phi-3.5 (Microsoft), DeepSeek-R1~\cite{deepseekr1}, and TinyLlama (Zhang et al.).

\end{document}